\documentclass[letterpaper]{article} 
\usepackage{aaai2026}
\usepackage{times}  
\usepackage{helvet}  
\usepackage{courier}  
\usepackage[hyphens]{url}  
\usepackage{graphicx} 
\usepackage{natbib}  
\usepackage{caption} 
\usepackage{algorithm}
\usepackage{algorithmic}

\usepackage{newfloat}
\usepackage{listings}
\DeclareCaptionStyle{ruled}{labelfont=normalfont,labelsep=colon,strut=off} 
\floatstyle{ruled}
\newfloat{listing}{tb}{lst}{}
\floatname{listing}{Listing}

\usepackage{booktabs}
\usepackage{makecell}
\usepackage{graphicx}
\usepackage{xcolor}
\usepackage{tcolorbox}
\tcbuselibrary{skins,breakable}
\usepackage[table]{xcolor}
\definecolor{accBest}{RGB}{198,239,206}   
\definecolor{accSecond}{RGB}{226,239,218} 
\definecolor{accThird}{RGB}{234,241,221}  
\usepackage{soul}

\newcommand{\hlbest}[1]{\sethlcolor{accBest}\hl{#1}}
\newcommand{\hlsecond}[1]{\sethlcolor{accSecond}\hl{#1}}
\newcommand{\hlthird}[1]{\sethlcolor{accThird}\hl{#1}}

\title{GamiBench: Evaluating Spatial Reasoning and 2D-to-3D Planning Capabilities of MLLMs with Origami Folding Tasks}

\author{
\begin{tabular}[t]{ccc}
    Ryan Spencer\textsuperscript{\rm 1}\thanks{Equal contribution.} & 
    Roey Yaari\textsuperscript{\rm 1}\footnotemark[1] & 
    Ritvik Vemavarapu\textsuperscript{\rm 1}\footnotemark[1] \\
    Joyce Yang\textsuperscript{\rm 1}\footnotemark[1] & 
    Steven Ngo\textsuperscript{\rm 1,2}\thanks{Correspondence.} & 
    Utkarsh Sharma\textsuperscript{\rm 1,3}\footnotemark[2]
\end{tabular}
}
\affiliations{
    \textsuperscript{\rm 1}Algoverse AI Research \quad
    \textsuperscript{\rm 2}UC San Diego \quad
    \textsuperscript{\rm 3}University of New South Wales\\
    
    svngo@ucsd.edu, utkarsh@algoverse.us
}

\begin{document}

\maketitle

\begin{abstract}

Multimodal large language models (MLLMs) are proficient in perception and instruction-following, but they still struggle with spatial reasoning: the ability to mentally track and manipulate objects across multiple views and over time. Spatial reasoning is a key component of human intelligence, but most existing benchmarks focus on static images or final outputs, failing to account for the sequential and viewpoint-dependent nature of this skill. To close this gap, we introduce \emph{GamiBench}, a benchmark designed to evaluate spatial reasoning and 2D-to-3D planning in MLLMs through origami-inspired folding tasks. \emph{GamiBench} includes 186 regular and 186 impossible 2D crease patterns paired with their corresponding 3D folded shapes, produced from six distinct viewpoints across three visual question-answering (VQA) tasks: predicting 3D fold configurations, distinguishing valid viewpoints, and detecting impossible patterns. Unlike previous benchmarks that assess only final predictions, \emph{GamiBench} holistically evaluates the entire reasoning process of the models; measuring cross-view consistency, physical feasibility through impossible-fold detection and interpretation of intermediate folding steps. It further introduces new diagnostic metrics—viewpoint consistency (VC) and impossible fold selection rate (IFSR)—to measure how well models handle folds of varying complexity. By linking geometric evaluation with sequential reasoning, \emph{GamiBench} enables a comprehensive evaluation of state-of-the-art MLLMs, revealing significant limitations in spatial reasoning capabilities, such as multi-view inconsistency and difficulty detecting physically impossible folds. Our experiments show that even leading models such as GPT-5 and Gemini-2.5-Pro struggle on single-step spatial understanding, while other MLLMs tend to show highly variable or inconsistent answering trends. These contributions establish a standardized framework for evaluating and advancing geometric understanding and spatial reasoning in MLLMs. The dataset and code is available at: \url{https://github.com/stvngo/GamiBench}.


\end{abstract}


\section{Introduction}

\begin{figure}[t]
    \centering
    \includegraphics[width=\linewidth, keepaspectratio]{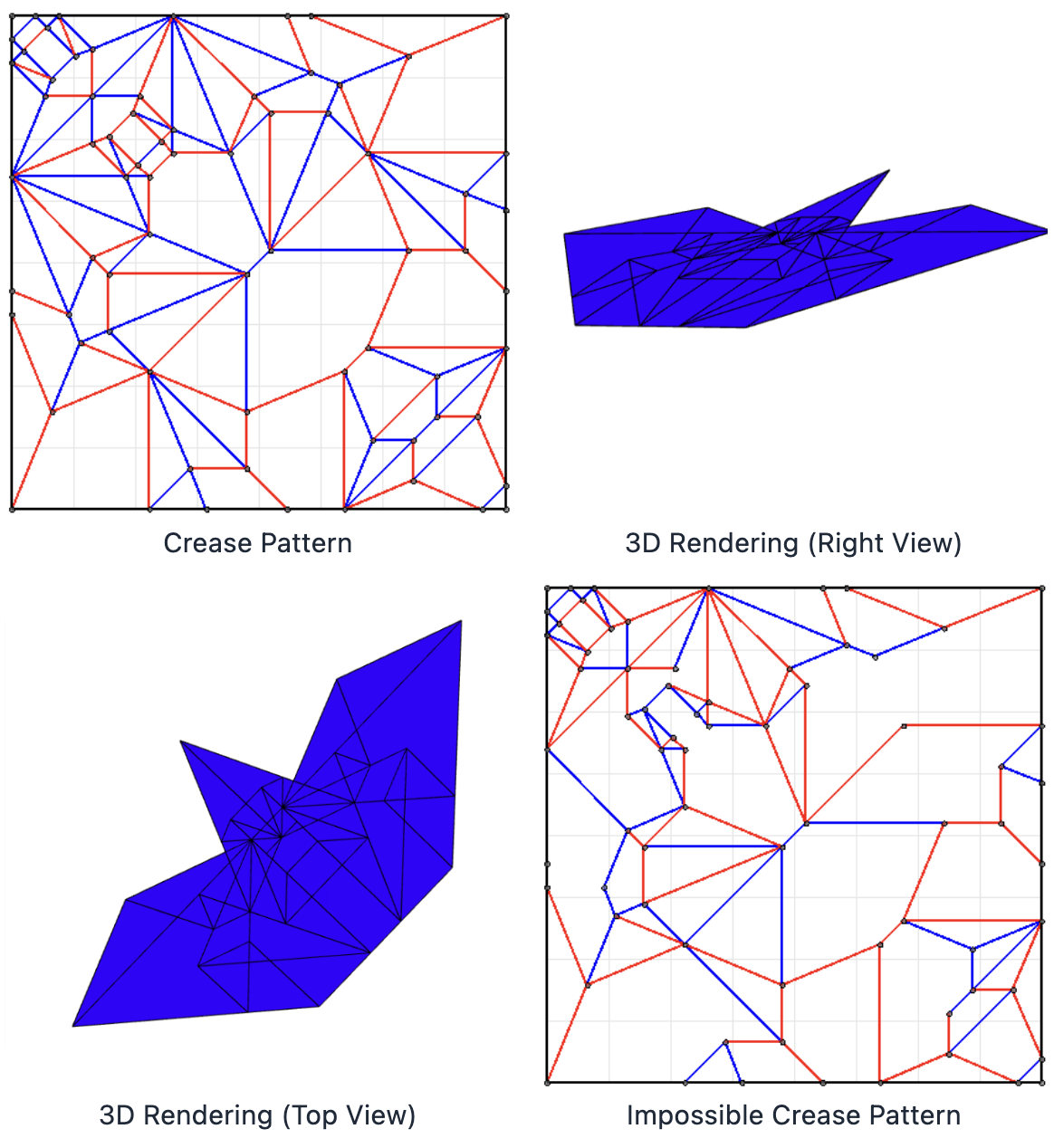}
    \caption{2D normal crease pattern (top-left), right-view 3D final fold state (top-right), top-view alternative 3D final fold state (bottom-left), and 2D impossible crease pattern (bottom-right) of an example butterfly variation (\textit{komatsu\_butterfly}). Impossible crease pattern has inverted creases and missing valley/mountain folds.}
    \label{fig:crease-pattern-example}
\end{figure}

Spatial reasoning is a fundamental component of human intelligence, crucial for interacting with the physical world, understanding relationships between objects, and executing multi-step actions. Tasks such as building furniture or folding origami require mentally simulating spatial transformations and tracking changing object states. As the interactions between artificial intelligence systems and real-world environments evolve, developing models that can reason about space and change has become a core challenge. Recent advances in multimodal large language models (MLLMs) show strong progress in image recognition, VQA, and instruction following \cite{dongfang2025omnidirectional}, \cite{jiang2025marble}; however, they clearly struggle on high-quality and temporally extended sequential spatial reasoning tasks \cite{rajabi2024gsrbench}, \cite{tang2025lego}, \cite{valmeekam2023planbench}.

Existing multimodal benchmarks and tools have advanced evaluation breadth (e.g., perception, captioning, instruction following) \cite{fu2023mme}, \cite{liu2023mmbench}, but typically emphasize single-step judgments, static scenes, or end-state accuracy \cite{fu2023mme}. Specialized efforts in spatial or procedural reasoning test important skills such as geometric consistency \cite{tang2025lego}, \cite{rajabi2024gsrbench}, arrangement from descriptions \cite{tang2025lego}, or plan generation \cite{valmeekam2023planbench}, but often isolate narrow capabilities, fix visual formats, or under-specify multi-view consistency and temporal coherence. Thus, currently there is no framework that holistically evaluates how models plan, update, and validate spatial states across time and across multiple viewpoints of the same 3D structure.

In this work, we introduce \emph{GamiBench}, a novel benchmark and framework for sequential spatial planning and reasoning that uses origami-inspired tasks to map 2D crease patterns to 3D multi-view final states (front, back, top, bottom, left, right). We take inspiration from the common traditional Japanese art of paper folding, Origami, using a single square sheet of paper to create a figure without cutting, gluing, or marking. The assembly of the 3D Origami model consists of dozens of intermediary discrete folds to transform a 2D paper plane into a 3D structure, providing an ideal foundation for testing sequential spatial reasoning abilities. We carefully curate a dataset of 186 regular and 186 impossible 2D square crease pattern configurations and 372 adjacent 3D folds, where each fold is part of a pair of correct folds for a given 2D crease pattern, using existing online tools like Oriedita \cite{oolbekkink2021oriedita} and Origami Simulator \cite{ghassaei2018origami}. \emph{GamiBench} encompasses a collection of 744 Visual Question-Answering (VQA) multiple-choice sets spanning across 3 tasks, one of which is conditioned on the final answer, from this dataset. 

First, we develop a set of tests to assess MLLMs’ basic spatial transformation understanding capabilities, including single-step 2D-to-3D mapping, reinforcement of initial correctness via alternative 3D final states, and assessment of impossible fold configuration acknowledgement. We develop our own metrics, such as Viewpoint Consistency (VC) and Impossible Fold Selection Rate (IFSR), to measure model performance across these 3 tasks.

\emph{GamiBench} provides several distinct advantages compared to existing spatial understanding benchmarks by evaluating (1) multi-perspective consistency across 3D views, (2) feasibility via impossible-fold detection (violations of physical origami axioms), and (3) sequential interpretation of intermediate states induced by textual instructions and visual transitions. Furthermore, we classify folds by complexity to stress-test models’ robustness to geometric density. 

Leveraging \emph{GamiBench}, we conduct evaluations on 21 state-of-the-art MLLMs, including proprietary models such as GPT-5 and Gemini-2.5-Flash, as well as leading open-source alternatives such as Llama-4-Maverick, Gemma-3-27B-IT, Cogito-V2-Preview-Llama-109B-MoE, and GLM-4.5V. Our results reveal a substantial gap in MLLM spatial reasoning proficiency. Even the strongest models struggle with basic spatial understanding tasks, at times performing worse than weaker models on the same task.

Concretely, our contributions are as follows:
\begin{itemize}
    \item A multi-view, sequential spatial benchmark that evaluates 2D-to-3D reasoning beyond final-state accuracy, with tasks that require coherence across six views and across time.
    \item New evaluation axes—Viewpoint Consistency and Impossible Fold Selection Rate (IFSR)—that diagnose failure modes missed by standard accuracy metrics.
    \item Origami-inspired task suite combining textual instructions with visual states to test whether MLLMs can plan, update, and verify geometric transformations.
    \item Complexity controls (simple vs. complex crease patterns) enabling analysis of scale effects on spatial planning.
\end{itemize}
Together, these contributions to spatial evaluation from static recognition toward procedural, multi-view reasoning provide clearer signals about where current MLLMs fall short and how future architectures or training regimes might improve.

\section{Related Work}
Various comprehensive benchmarks have been introduced to assess various multimodal capabilities. 

\textbf{Foundations of Spatial and Embodied Reasoning.} Spatial intelligence has long been identified as a fundamental component of human cognition, underlying reasoning about geometry and physical relationships \cite{bornstein1986psychology}. Recent embodied MLLM frameworks extend this concept to machine perception and control. OpenVLA \cite{kim2024openvla} links vision, language, and action to achieve generalist visuomotor manipulation across diverse robots, while ManipLLM \cite{li2024manipllm} integrates affordance reasoning and pose prediction for object-centric manipulation in real environments. In autonomous driving, DriveMLLM \cite{guo2024drimellm} benchmarks spatial scene understanding under occlusion and dynamic layouts, and DriveMLM \cite{wang2023drivemlm} aligns multimodal perception with behavioral-planning states for closed-loop navigation. Follow-up studies reveal that modality imbalance limits generalization from simple to complex visual reasoning tasks \cite{park2025generalizing}, and that MLLMs still struggle to remember and reconstruct 3D spaces from sequential observations \cite{yang2024thinking}. Collectively, these works highlight the gap between embodied perception and sustained geometric reasoning across changing viewpoints.

\textbf{General Multimodal Evaluation Benchmarks.} Comprehensive multimodal benchmarks have advanced large-scale evaluation of perception and reasoning. MME \cite{fu2023mme} systematically measures 14 multimodal subtasks, exposing persistent weaknesses such as object hallucination and spatial reasoning errors. MMBench \cite{liu2023mmbench} introduces a bilingual multiple-choice framework for fine-grained multimodal assessment, while MMMU \cite{yue2024mmmu} extends difficulty to college-level, discipline-specific visual reasoning. Despite their breadth, these datasets primarily test static understanding and lack mechanisms for evaluating multi-view geometric coherence or physical feasibility, both of which are core aspects addressed by \emph{GamiBench}.

\textbf{Spatial and Geometric Reasoning.} Recent benchmarks directly target spatial reasoning capabilities. GSR-Bench \cite{rajabi2024gsrbench} evaluates object-relation understanding and shows that MLLMs frequently confuse depth and relative position, revealing weak geometric grounding. LEGO-Puzzles \cite{tang2025lego} probes multi-step spatial reasoning through LEGO-based assembly tasks and uncovers severe performance gaps between humans ($\sim$90\%) and MLLMs ($\sim$50\%). 3DSRBench \cite{ma2024threedsrbench} assesses 3D reasoning across orientation, occlusion, and viewpoint changes, finding that accuracy drops sharply under non-canonical perspectives. OSR-Bench \cite{dongfang2025omnidirectional} examines omnidirectional spatial reasoning over 360° panoramic inputs and reports poor rotation invariance in current models. Psychometric analysis of basic spatial abilities \cite{xu2025spatialabilities} corroborates these findings, identifying deficits in mental rotation and coordinate transformation. Zhang et al. \cite{zhang2025call} argue that such weaknesses cannot be solved through scaling alone and call for geometry-aware, physically grounded training data. \emph{GamiBench} complements these efforts by uniting physical validity and cross-view spatial consistency through origami-inspired folds.

\textbf{Sequential Planning and Temporal Reasoning.} Beyond static geometry, MARBLE \cite{jiang2025marble} tests multimodal reasoning and planning across multi-step spatial tasks, showing near-random performance even in simplified subtasks. PlanBench \cite{valmeekam2023planbench} provides a textual framework for evaluating reasoning about actions and change, serving as a foundation for later multimodal planning benchmarks. Similarly, \emph{GamiBench} models sequential folding as a structured planning problem, requiring consistency across transformations and temporal stages.

\textbf{Interpretability and Evaluation Bias.} VERIFY \cite{bi2025verify} isolates visual reasoning fidelity using human-annotated reasoning paths, revealing that models often reach correct answers for the wrong visual evidence. In parallel, Zheng et al. \cite{zheng2024robust} expose a systemic multiple-choice selection bias in LLMs, showing preference for specific option positions independent of content. \emph{GamiBench} incorporates an Impossible Fold Selection Rate (IFSR) metric and balanced MCQ design to mitigate such biases, ensuring that geometric reasoning rather than positional heuristics drives model performance.

Collectively, these studies have broadened multimodal reasoning evaluation yet remain restricted to static perception, symbolic planning, or 2D scene understanding. None directly assess 2D-to-3D transformation, multi-view spatial coherence, or physical feasibility. \emph{GamiBench} addresses this gap through origami-inspired tasks that couple 2D crease patterns with multi-view 3D structures, employing Viewpoint Consistency (VC) and Impossible Fold Selection Rate (IFSR) to capture how MLLMs reason about geometry, transformation, and physical constraints.

\section{Methodology}
\emph{GamiBench} is a multimodal benchmark that evaluates MLLMs on two-dimensional-to-three-dimensional spatial reasoning through origami-inspired tasks. Designed to evaluate both perception and procedural planning, \emph{GamiBench} formalizes spatial reasoning as the process of mapping crease patterns to structures while maintaining geometric consistency throughout multiple viewpoints.

\begin{figure}[t]
    \centering
    \includegraphics[width=\linewidth, keepaspectratio]{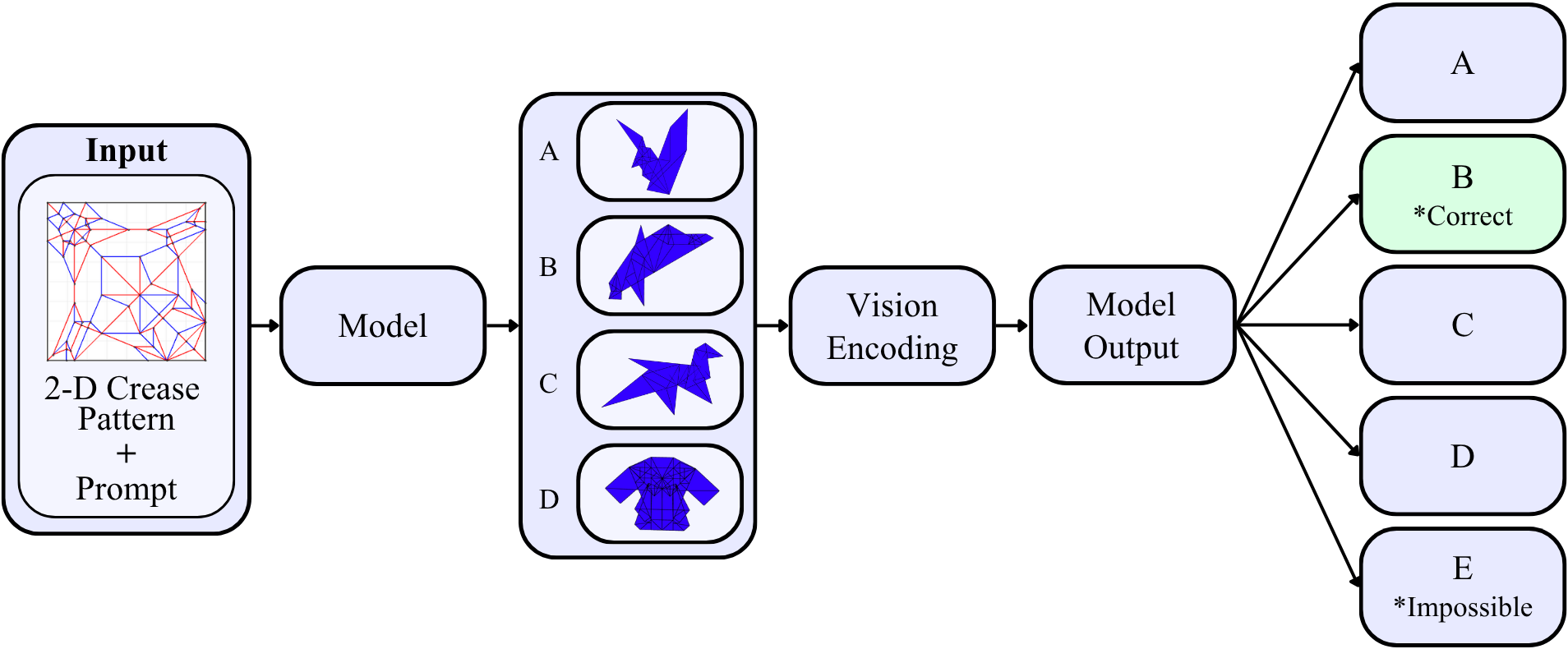}
    \caption{Visual mapping of the task flow in GamiBench. The model receives a 2D crease pattern (\emph{komatsu\_dolphin}) and text prompt, encodes candidate 3D folds, and outputs the most plausible match among multiple choices.}
    \label{fig:visual-mapping}
\end{figure}

\subsection{Task Definition}
\emph{GamiBench} makes use of two task clusters showing differing reasoning regimes: (i) Single-Step Spatial Understanding (SSSU) that measures an MLLM’s ability to infer a folded 3D structure based on a single 2D crease pattern, and (ii) Multi-Step Spatial Reasoning that requires temporal reasoning within the context of cross-view coherence.

For the one-step setup, models will produce final shape identification and view point recognition. The multi-step setting will extend this reasoning over to complex folding progressions, requiring models to be cognizant of geometric transitions and determine the step at which a fold can or cannot feasibly take place; thus exposing the models understanding of spatial continuity and physical possibility. This multi-step setting is \emph{implicit}; we do not prompt models with intermediate folding steps. Our multi-step process tests models' multi-step understanding of complex folds with 40+ creases that require lots of spatial transformations. An overview of the task formulation and model decision flow is shown in Figure \ref{fig:visual-mapping}. For analysis of scaling behavior, folds are classified into two levels; simple folds (less than 40 creases), and complex folds (greater than or equal to 40 creases).

\subsection{Dataset Curation}


\textbf{Dataset Collection.} The dataset comprises 186 origami instances, each containing a normalized 2D crease pattern, 2D impossible crease pattern, and two 3D renderings captured from canonical viewpoints. For each instance, only two of the six canonical viewpoints—front, back, top, bottom, left, and right—are selected and verified by humans for plausibility. To construct these instances, we sourced crease patterns from the Flat-Folder platform \cite{ku2025flatfolder}, which provides publicly available flat-foldable crease configurations, and verified their geometry using the open-source software Oriedita \cite{oolbekkink2021oriedita}, which enables precise control over mountain–valley assignments and fold geometry. Each pattern was then simulated and exported as a 3D folded mesh via Origami Simulator \cite{ghassaei2018origami}. All meshes were rendered under identical lighting and camera conditions to ensure viewpoint consistency. We include both physically valid and constraint-violating (impossible) folds to evaluate model sensitivity to geometric feasibility. Each instance was normalized and aligned to a shared coordinate frame to maintain comparability across samples.

\textbf{Foldability Verification.} We verify the feasibility of each crease pattern using Oriedita’s built-in programmatic verifier, CAMv \cite{oolbekkink2021oriedita}, which automatically applies flat-foldability constraints derived from origami axioms and theorems such as Kawasaki’s \cite{barile2002kawasaki} and Maekawa’s \cite{maekawa1983theorem}. This ensures that physically valid folds satisfy all geometric consistency conditions, while infeasible folds violate at least one constraint. The verifier’s outputs are cross-checked during data generation to confirm that impossible folds arise from genuine geometric contradictions rather than rendering artifacts.

\begin{figure}[t]
\centering
\fontsize{9}{10}\selectfont
\begin{tcolorbox}[
  enhanced, width=\columnwidth, 
  colback=gray!5, colframe=gray!50,
  boxrule=0.6pt, arc=1mm,
  left=2mm, right=2mm, top=1mm, bottom=1mm,
  title={\textbf{Task-Specific Template (here for task \textit{SSSU})}},
  colbacktitle=gray!70!black, coltitle=white,
  fonttitle=\bfseries, attach boxed title to top center={yshift=-2pt},
  boxed title style={sharp corners, frame hidden},
  halign=flush left, 
  before upper=\setlength{\parindent}{0pt}\raggedright
]
\textbf{Instruction:} 
You are an \textbf{Origami Folding Expert}. You will be given the final crease pattern of a folded origami model and four candidate 3D models labeled A through D. Evaluate \textbf{all four} symmetrically—do not privilege any order. Only one of the candidate models corresponds exactly to the result of folding the given crease pattern. In the crease pattern, \textcolor{red}{red} lines represent \emph{mountain} folds and \textcolor{blue}{blue} lines represent \emph{valley} folds. Your task is to analyze the crease pattern and select the correct 3D model based solely on visual and geometric reasoning. Consider fold types, symmetry, flap orientation, and structural features visible in the crease pattern. If none of the four models are possible, respond with option \textbf{E} (“This fold is impossible”). At each stage, respond with a single uppercase letter \textbf{A}, \textbf{B}, \textbf{C}, \textbf{D}, or \textbf{E}.

\medskip
\textbf{Question:} 
What is the correct 3D model for the given 2D crease pattern?

\medskip
\textbf{Answer:} 
Ground Truth
\end{tcolorbox}

\caption{Task-specific template. Our QA template includes instructions, question, and answer
for the \emph{SSSU} task.}
\label{fig:origami-template}
\end{figure}

\textbf{Origami Axioms.} To distinguish physically valid folds from infeasible ones, we define fold legality according to geometric and physical origami constraints. A fold is considered feasible if it satisfies flat-foldability conditions such as Kawasaki’s \cite{barile2002kawasaki} and Maekawa’s theorems \cite{maekawa1983theorem} and maintains continuous paper geometry without self-intersection. Impossible folds are generated by deliberately violating these principles, for example, by enforcing inconsistent mountain–valley assignments or overlapping crease intersections that would cause self-intersections in a real sheet. Through this structured design, \emph{GamiBench} will test MLLMs understanding of spatial transformations and ability to maintain consistency across multi-view geometric representations. 
For completeness, we include the Huzita--Hatori axioms \citep{huzita1991axioms, hatori2002axioms}, which define the geometric foundations of origami foldability. These axioms state that a fold can pass through any two given points, or place one point onto another; that a fold can align two lines or pass through a point while being perpendicular to a given line; and that more complex conditions allow folds to place one point onto a line while passing through another point, to map two points each onto their respective lines, or to position a point onto one line while remaining perpendicular to another. Collectively, these seven axioms describe the complete set of single-fold operations possible under Euclidean geometry and serve as the basis for distinguishing valid from infeasible fold structures in our dataset.

\textbf{Question-Answer Generation and Quality Control.} To streamline our evaluations of MLLMs, we design a template for question-answer generation. See Figure \ref{fig:origami-template} for an example. Each data example includes a multimodal instruction, multiple-choice prompt, and an answer. For each 2D data example, we define its 3D fold label as the correct answer, randomly assigned to a letter between A-D. Additionally, we randomly sample 3 other 3D viewpoints in our dataset as incorrect answers, assigned to the remaining letters to build our multiple-choice answer bank. Our prompt remains immutable throughout all 3 tasks (See Figure \ref{fig:origami-template}). We condition the viewpoint consistency task on the event that the standard task is correctly answered.

To maintain reproducibility and minimize duplication errors, five human annotators reviewed each multiple-choice set, verifying (1) the absence of duplicated or mirrored folds, (2) the balanced difficulty across options, and (3) the lack of visually ambiguous distractors. These checks ensured that the final answer banks remained balanced, non-redundant, and free from biases that could influence model responses. We also set a seed of 42 for randomizing answer choices and a temperature of 0 for most models to facilitate deterministic outputs.

\section{Results}

\begin{table*}[t]
\centering
\setlength{\tabcolsep}{3pt}
\renewcommand{\arraystretch}{1.12}
\begin{tabular}{lcccccccccc}
\toprule
& \multicolumn{4}{c}{\textbf{Simple}} & \multicolumn{4}{c}{\textbf{Complex}} & \multicolumn{2}{c}{\textbf{Overall}} \\
\cmidrule(lr){2-5}\cmidrule(lr){6-9}\cmidrule(lr){10-11}
\textbf{Model} & Normal & VC & IFSR & Imp. &
                 Normal & VC & IFSR & Imp. &
                 Normal & Imp. \\
\midrule
\multicolumn{11}{l}{\textit{Closed Source}}\\
Claude Opus 4.1        & 44.9 & 84.6 & 22.2 & 14.3 & 38.7 & 84.2 & 22.4 & 11.4 & 41.8 & 12.9 \\
Claude Opus 4          & 38.8 & 63.2 & 34.1 & 12.2 & 36.5 & 64.1 & 34.9 & 15.9 & 37.7 & \cellcolor{accThird}14.1 \\
Claude 4.5 Sonnet      & 34.7 & 82.0 & 20.6 &  0.0 & 48.2 & 88.2 & 19.0 & 16.7 & 41.5 &  8.4 \\
Grok-4-Fast            & 36.7 & 70.7 & 29.4 &  0.0 & 33.6 & 64.0 & 31.7 &  6.1 & 35.2 &  3.1 \\
GPT-o3                 & 44.9 & 80.7 & 26.2 &  8.2 & 38.0 & 81.9 & 25.4 &  9.8 & 41.5 &  9.0 \\
GPT-5                  & 67.3 & 88.4 & 25.4 &  6.1 & 60.6 & 88.9 & 24.6 & 10.6 & \cellcolor{accBest}64.0 &  8.4 \\
GPT-4o                 & 61.2 & 83.9 & 27.0 &  8.2 & 38.7 & 82.2 & 28.6 & 18.2 &  \cellcolor{accSecond}50.0 & 13.2 \\
GPT-4o-Mini            & 26.5 & 84.6 & 22.2 & 22.4 & 34.3 & 85.7 & 21.4 & 34.1 & 30.4 & \cellcolor{accBest}28.3 \\
Gemini-2.0-Flash       & 26.5 & 83.5 & 22.2 & 10.2 & 29.9 & 85.0 & 23.0 & 25.8 & 28.2 & \cellcolor{accSecond}18.0 \\
Gemini-2.5-Pro         & 49.0 & 90.3 & 15.9 &  8.2 & 43.8 & 91.2 & 15.1 & 12.1 & \cellcolor{accThird}46.4 & 10.2 \\
Gemini-2.5-Flash       & 49.0 & 88.3 & 17.5 &  8.2 & 38.0 & 89.5 & 16.7 &  5.3 & 43.5 &  6.8 \\
\midrule
\multicolumn{11}{l}{\textit{Open Source}}\\
Mistral-Medium-3.1     & 26.5 & 84.6 & 22.2 & 42.0 & 24.8 & 82.5 & 24.6 & 40.2 & 25.7 & \cellcolor{accBest}41.1 \\
Llama-4--Scout         & 34.7 & 64.1 & 26.2 &  0.0 & 44.5 & 67.4 & 26.2 &  0.0 & 39.6 &  0.0 \\
Llama-4--Maverick      & 24.5 & 59.2 & 23.0 &  0.0 & 43.8 & 59.0 & 29.4 &  0.8 & 34.2 &  0.4 \\
Gemma-3--27B-IT        & 38.8 & 86.2 & 23.0 & 12.0 & 46.0 & 86.0 & 22.2 &  8.3 & \cellcolor{accThird}42.4 & \cellcolor{accThird}10.2 \\
\makecell[l]{Cogito-V2-Preview\\Llama-109B-MoE} & 49.0 & 88.1 & 22.2 &  6.0 & 39.4 & 78.3 & 22.2 & 15.9 & \cellcolor{accBest}44.2 & \cellcolor{accSecond}11.0 \\
Qwen3-VL-8B--Thinking  & 28.6 & 67.6 & 22.2 &  0.0 & 27.0 & 64.0 & 23.0 &  0.0 & 27.8 &  0.0 \\
Qwen3-VL-30B-A3B--Thinking & 26.5 & 66.1 & 24.6 &  0.0 & 32.8 & 69.0 & 24.6 &  0.8 & 29.7 &  0.4 \\
Qwen3-VL-235B-A22B--Thinking & 24.5 & 66.7 & 22.2 &  2.0 & 35.8 & 72.9 & 22.2 &  0.8 & 30.2 &  1.4 \\
Microsoft Phi-4 Multimodal-Instruct & 36.7 & 66.7 & 27.0 &  2.0 & 38.7 & 72.9 & 23.8 &  3.0 & 37.7 &  2.5 \\
GLM-4.5V               & 46.9 & 78.3 & 23.0 &  6.0 & 40.1 & 79.4 & 23.8 &  4.5 & \cellcolor{accSecond}43.5 &  5.3 \\
\bottomrule
\end{tabular}

\caption{GamiBench results (percent). Two-level headers: Simple vs. Complex; rightmost columns macro-average across both complexities. VC = viewpoint consistency. Imp. = Impossible. Overall \hlbest{best}, \hlsecond{second best}, and \hlthird{third best} are highlighted as such.}
\label{tab:gami_combined_overall_aaai}
\end{table*}

\subsection{Overall Performance}
Among the 21 MLLMs evaluated in \emph{GamiBench}, performance varied significantly, particularly when transitioning from simple to complex folding tasks (See Table \ref{tab:gami_combined_overall_aaai}). Most models demonstrated reasonable competence in SSSU, such as interpreting a 2D crease pattern and identifying its corresponding 3D structure. However, accuracy declined sharply when multi-step reasoning was required. Tasks that demanded tracking fold sequences or maintaining geometric consistency across temporal stages were especially challenging, revealing limitations in the model's ability to integrate spatial transformations over time.

\textbf{Evaluation Metrics.} To ensure consistent assessment across all three VQA tasks, \emph{GamiBench} employs three complementary measurements: Accuracy, Impossible Fold Selection Rate (IFSR), and Viewpoint Consistency (VC). Accuracy captures the model’s ability to identify the correct 3D fold or viewpoint among four multiple choice options, providing a direct measure of categorical prediction quality. IFSR quantifies how often a model incorrectly labels a valid fold as impossible, reflecting its sensitivity to geometric infeasibility and physical constraint reasoning. VC measures whether a model that correctly identifies the 3D fold for a crease pattern from one viewpoint remains correct when the same 3D fold is presented from a different viewpoint (randomly chosen from the remaining available views), using the same 3 distractor candidates as in the primary trial. Thus, VC here is a \textit{conditional} single-view re-test accuracy: the fraction of eligible items whose correctness persists under a viewpoint change with an unchanged candidate set except for the new correct image (the denominator is the number of primary successes). High VC scores indicate consistent multi-view understanding, while lower scores reveal discrepancies in spatial alignment or rotation tracking.






Let $\mathcal{D}$ be the set of normal crease patterns with a valid 3D fold.
For item $i \in \mathcal{D}$, let $y_i$ be the correct 3D fold, $v_0$ the
primary view, and $v_1 \neq v_0$ the re-test view (same three distractors).

\textbf{Normal accuracy.}
\begin{equation}
\mathrm{Acc} \;=\; \frac{1}{|\mathcal{D}|} \sum_{i \in \mathcal{D}}
\mathbf{1}\!\left[\, \hat{y}_i^{(v_0)} \,=\, y_i \,\right]
\end{equation}

Define the subset of primary-view successes
\begin{equation}
\mathcal{S} \;=\; \bigl\{\, i \in \mathcal{D} \,:\, \hat{y}_i^{(v_0)} = y_i \,\bigr\}.
\end{equation}

\textbf{Conditional viewpoint consistency.}
\begin{equation}
\mathrm{VC} \;=\; \frac{1}{|\mathcal{S}|} \sum_{i \in \mathcal{S}}
\mathbf{1}\!\left[\, \hat{y}_i^{(v_1)} \,=\, y_i \,\right]
\end{equation}



\subsection{Closed-Source Models}
\begin{figure}
    \centering
    \includegraphics[width=\linewidth, keepaspectratio]{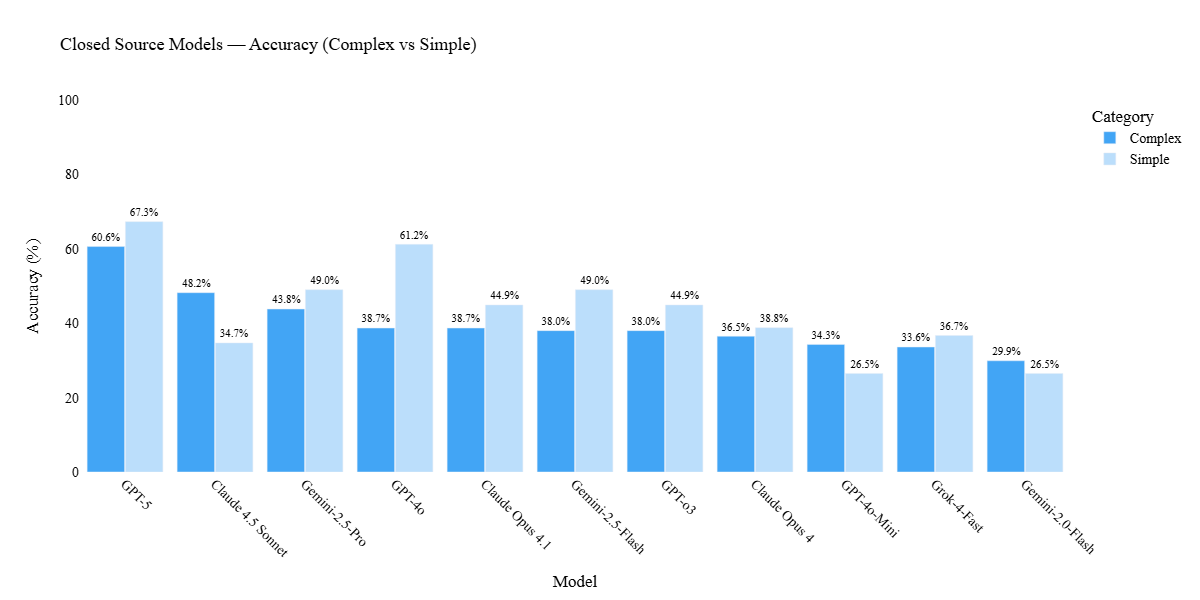}
    \includegraphics[width=\linewidth, keepaspectratio]{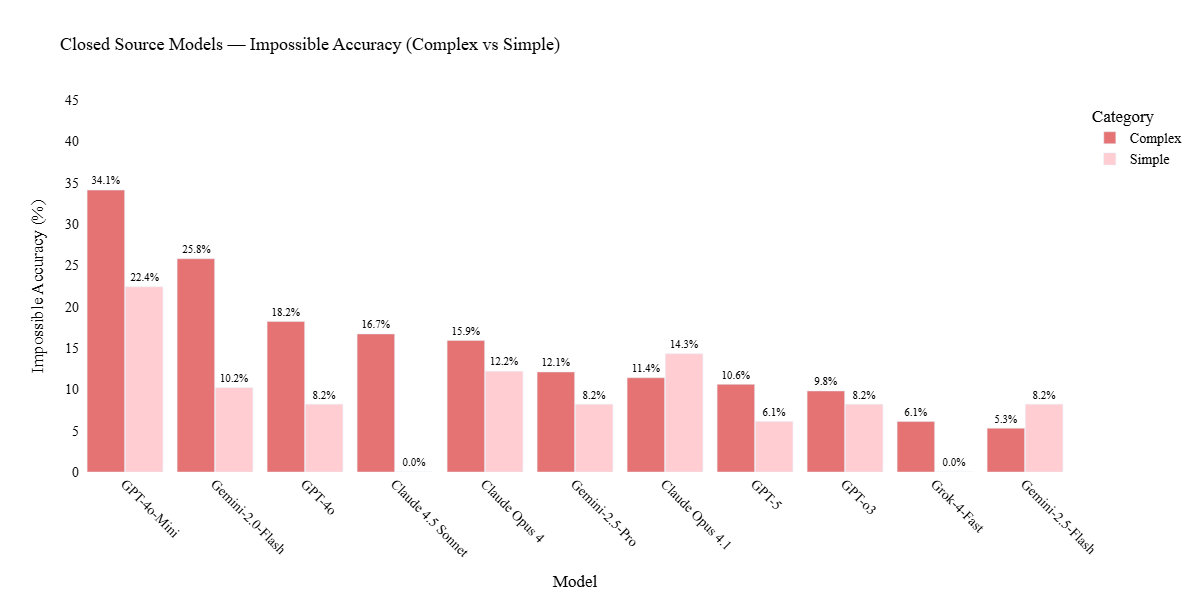}
    \caption{GamiBench Closed-Source Evaluations. Regular accuracy (top) and Impossible accuracy (bottom) of models, sorted in descending order from left to right, complex and simple.}
    \label{fig:gami-closed-source-graphs}
\end{figure}

Among closed-source systems, GPT-5 stood out as the strongest performer. It correctly identified 60.6\% of complex folds and 67.3\% of simple folds, while maintaining relatively high viewpoint consistency and angular stability (VC: 65.1\% and 69.7\%). These results suggest that GPT-5 has a comparatively stable understanding of 3D spatial relationships across multiple views.

In contrast, Grok-4-Fast and Gemini-2.0-Flash were the least effective in this group, with complex-fold accuracies of 33.6\% and 29.9\%, respectively. Claude Opus 4.1 and Claude 4.5 Sonnet achieved midrange results, with complex-fold accuracies between 38.7\% and 48.2\%, though the latter model showed stronger spatial coherence (VC up to 82.4\%). GPT-4o performed well on simpler folds (61.2\%) but plateaued on complex ones (38.7\%), a trend consistent across most closed-source models.

Across this group, the Impossible Fold Selection Rate (IFSR), which measures how often a model incorrectly classifies a valid fold as impossible, remained consistently low, typically between 0–14\%. Although a low rate might initially appear favorable, it actually reveals a limited sensitivity to geometric infeasibility: models rarely identify impossible folds even when they should. This suggests that despite strong visual reasoning, most closed-source MLLMs still lack a robust understanding of physical constraints and spatial validity.

\subsection{Open-Source Models}
\begin{figure}
    \centering
    \includegraphics[width=\linewidth, keepaspectratio]{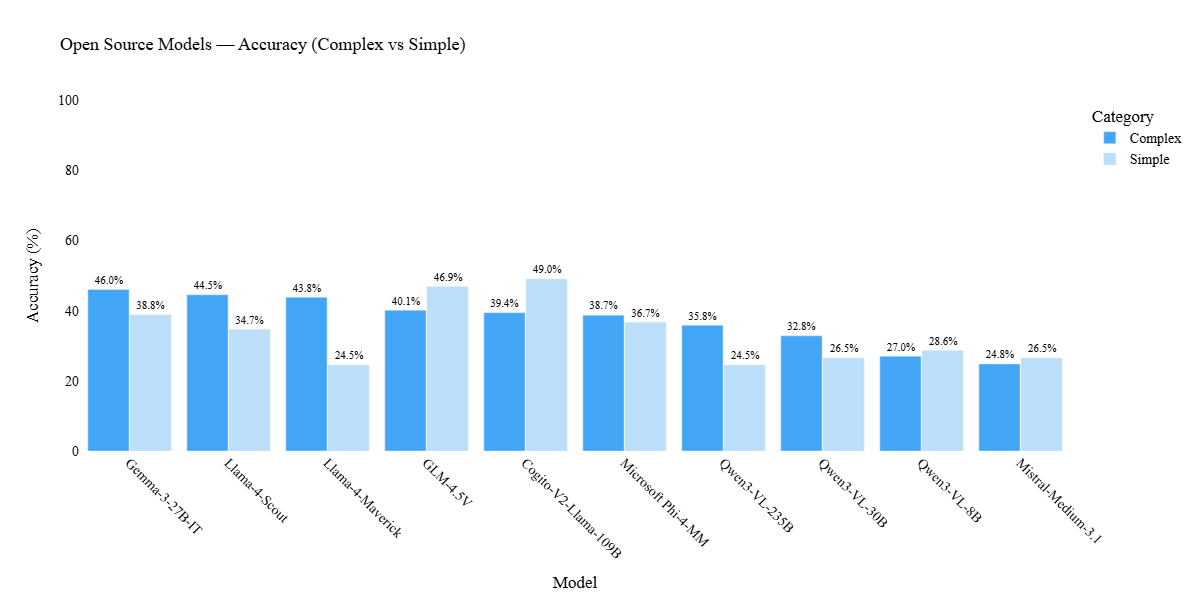}
    \includegraphics[width=\linewidth, keepaspectratio]{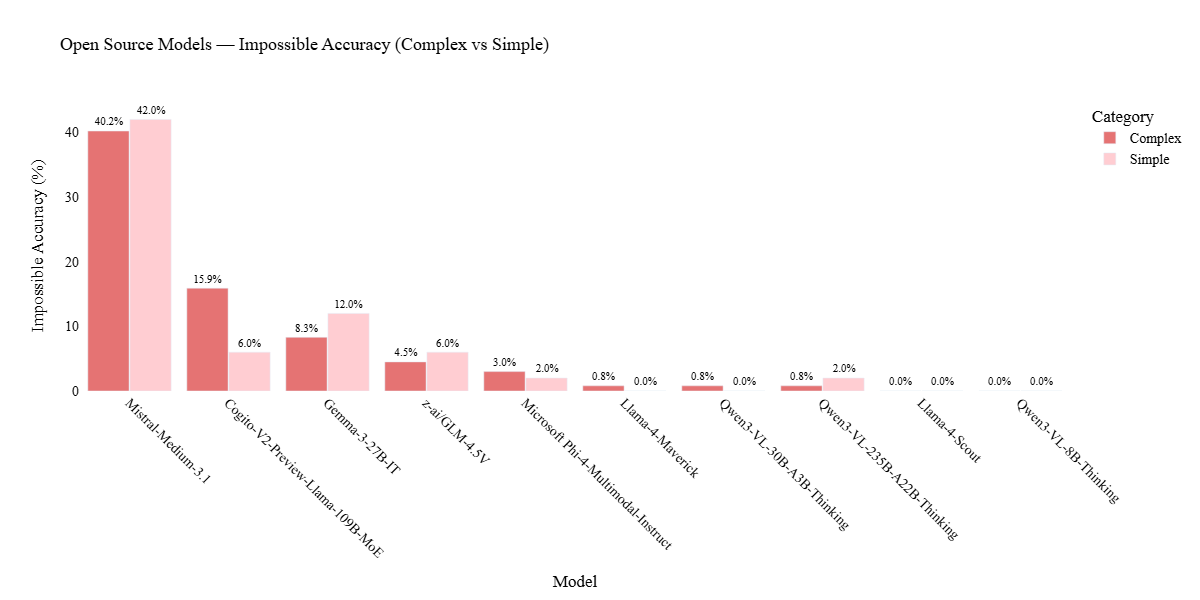}
    \caption{GamiBench Open-Source Evaluations. Regular accuracy (top) and Impossible accuracy (bottom) of models, sorted in descending order from left to right, complex and simple.}
    \label{fig:open-source-graphs}
\end{figure}

Open-source models exhibited wider variability but often performed surprisingly well relative to their commercial counterparts. Llama-4-Scout, Llama-4-Maverick, and Gemma-3-27B-IT reached complex-fold accuracies between 44–46\%, matching or slightly exceeding several closed-source systems. More impressively, their VC scores frequently surpassed 80\%, suggesting a strong ability to maintain visual and geometric consistency across viewpoints, even when their final predictions were not always correct.

Cogito-V2-Preview-Llama-109B-MoE balanced accuracy and coherence particularly well, achieving 39.4\% on complex folds and 49.0\% on simple folds, with outstanding VC performance (85–92\%). On the other hand, the Qwen3-VL models (8B, 30B, and 235B) consistently underperformed, recording accuracies below 33\% on complex folds.

The GLM-4.5V model emerged as one of the most capable open systems, achieving 40.1\% accuracy on complex folds and 46.9\% on simple ones. Its VC scores (87.3\% and 78.3\%) placed it among the most consistent models in cross-view reasoning, showing that open models can indeed compete on structural coherence even without proprietary optimization.

\subsection{Trends and Observations}
Across all systems, task complexity was the clearest predictor of performance. On average, models performed 10–15\% better on simple folds than on complex ones. Complexity controls in \emph{GamiBench} categorize folds according to geometric density and reasoning depth. For our purposes, simple folds (fewer than 40 creases) primarily test localized spatial mapping and single-step 2D-to-3D transformations. In contrast, complex folds (40 or more creases) require multi-step reasoning, constraint tracking, and cross-view geometric coherence. Interpreting our results through this framework shows that increasing geometric density substantially amplifies cognitive and planning demands. While most MLLMs maintain stable accuracy on simple folds, their performance degrades as structural interactions grow combinatorially. This trend validates \emph{GamiBench}’s complexity controls as a diagnostic tool for assessing how model reasoning scales with geometric and procedural difficulty. However, higher viewpoint consistency did not always imply correctness. Some models produced geometrically consistent yet incorrect shapes, revealing what we term a visual plausibility bias. In such cases, the models could maintain coherent visuals while misunderstanding the underlying 3D geometry.

Another persistent limitation was the inability to distinguish between valid and impossible folds. Even top-performing models like Mistral-Medium-3.1 and GPT-4o-Mini achieved combined average IFSRs of 27.6\%, and 14.8\%, respectively. This confirms that reasoning about spatial constraints and physical feasibility remains an open challenge in multimodal modeling.

We observe a miscalibration in Llama-4–Scout and Llama-4–Maverick: both select “impossible” 0\% of the time on truly impossible items, yet show IFSR = 26.2\%, indicating they sometimes (incorrectly) choose “impossible” on normal items. This asymmetric error pattern reflects a decision bias against the “E” option under ground-truth impossible conditions, limiting reliability on feasibility detection.

Finally, we observed a subtle but consistent selection bias in multiple-choice evaluations. Several models showed a tendency to favor specific answer options (e.g., “Option A”) regardless of content, echoing patterns noted in prior work \cite{zheng2024robust}. Future benchmark iterations should further mitigate this issue by incorporating generative, open-response formats that reduce residual positional bias and more effectively capture authentic reasoning ability. 

\subsection{Limitations}
While \emph{GamiBench} provides a structured and interpretable framework for assessing 2D-to-3D spatial reasoning, several limitations remain. The benchmark focuses on synthetic origami-inspired folds and does not yet account for real-world conditions such as material deformation, lighting variation, or visual clutter, which may limit its generalization beyond crease-pattern reasoning. We also do not conduct \emph{true} multi-step tasks such as predicting intermediary folds between 2D and 3D. The dataset size is moderate, and its complexity definition, which relies primarily on crease count, may not accurately reflect true planning difficulty caused by symmetry or long-range geometric dependencies. This is further supported by our finding that some models performed better on complex folds than on simple ones (i.e. Llama-4 series, Claude 4.5 Sonnet), indicating that crease count alone does not fully capture the true planning difficulty.

Furthermore, evaluations were performed under a single prompt and decoding configuration without systematic temperature tuning or standardized API normalization. Because proprietary APIs change over time, reproducibility remains partially limited. Future work could establish a human baseline to contextualize model performance, increase data realism and scale, introduce interactive folding and feedback-based tasks, integrate calibrated uncertainty metrics, and standardize evaluation protocols to improve comparability across evolving MLLM architectures.

\section{Conclusion}
Thus, we introduce \emph{GamiBench}, a new benchmark for evaluating spatial reasoning and 2D-to-3D planning that aims to push the limits of modern MLLMs. By coupling origami-inspired crease patterns with multi-view 3D states, \emph{GamiBench} advances beyond static visual understanding toward dynamic and multi-perspective spatial reasoning. Our results show that although leading MLLMs exhibit emerging competence in single-step geometric inference, they struggle to maintain spatial coherence across time, viewpoints, and physical constraints. Looking ahead, we envision \emph{GamiBench} as a foundation for future spatial reasoning research, encouraging the development of models that can truly think in space by integrating perception, geometry, and physical reasoning into a unified understanding of real-world dynamics.

\bibliography{aaai2026}

\end{document}